\documentclass[letterpaper, 10 pt, conference]{ieeeconf}  

\IEEEoverridecommandlockouts                              

\usepackage{amsmath} 
\usepackage{amssymb}  
\usepackage{graphics}
\usepackage{xspace}

\usepackage{amsmath}
\usepackage{amssymb}
\usepackage{mathtools}

\usepackage{hyperref}
\usepackage{url}

\usepackage[utf8]{inputenc} 
\usepackage[T1]{fontenc}    

\usepackage{scalefnt,letltxmacro}
\LetLtxMacro{\oldtextsc}{\textsc}
\renewcommand{\textsc}[1]{\oldtextsc{\scalefont{1.10}#1}}

\usepackage{caption}
\usepackage{subcaption}

\usepackage{microtype}
\usepackage{amsfonts,amsmath,amssymb}   
\usepackage{graphicx}         
\usepackage{booktabs,tabularx}          
\usepackage{nicefrac}                   
\usepackage{graphics}                   
\usepackage{multirow}
\usepackage{lipsum}  
\usepackage[font=footnotesize]{caption}
\usepackage{adjustbox}
\usepackage{soul}
\usepackage{xcolor}
\usepackage{pdfpages}
\usepackage{listings}
\usepackage{xspace}
\usepackage{wrapfig}
\usepackage{cite}
\usepackage{xcolor}

\definecolor{mygreen}{rgb}{0.1, 0.6, 0.1}

\usepackage{titlesec}
\titlespacing\section{0pt}{4pt plus 2pt minus 2pt}{2pt plus 2pt minus 2pt}
\titlespacing\subsection{0pt}{4pt plus 4pt minus 2pt}{2pt plus 2pt minus 2pt}
\titlespacing\subsubsection{0pt}{3pt plus 4pt minus 2pt}{0pt plus 2pt minus 2pt}

\definecolor{Blue9}{rgb}{0.098,0.3,0.9}
\definecolor{DarkBlue}{rgb}{0,0.08,0.45}
\usepackage{hyperref}
\hypersetup{
  colorlinks,
  urlcolor  = Blue9,
  linkcolor = Blue9,
  citecolor = Blue9,
}

\newcommand{\ours}{PIP\xspace}

\title{\LARGE \bf
Exploiting Policy Idling for Dexterous Manipulation
}

\author{Annie S. Chen$^1$\thanks{$^1$Stanford University work done while interning at Google DeepMind}, Philemon Brakel$^2$\thanks{$^2$Google DeepMind}, Antonia Bronars$^3$\thanks{$^3$MIT work done while interning at Google DeepMind}, Annie Xie$^2$, Sandy Huang, Oliver Groth$^2$, \\ Maria Bauza$^2$, Markus Wulfmeier$^2$, Nicolas Heess$^2$, Dushyant Rao$^2$ \\
}

\begin{document}

\maketitle
\thispagestyle{empty}
\pagestyle{empty}

\begin{abstract}

Learning-based methods for dexterous manipulation have made notable progress in recent years. However, learned policies often still lack reliability and exhibit limited robustness to important factors of variation. One failure pattern that can be observed across many settings is that policies idle, i.e. they cease to move beyond a small region of states when they reach certain states. This \emph{policy idling} is often a reflection of the training data. For instance, it can occur when the data contains small actions in areas where the robot needs to perform high-precision motions, e.g., when preparing to grasp an object or object insertion.
Prior works have tried to mitigate this phenomenon e.g. by filtering the training data or modifying the control frequency. However, these approaches can negatively impact policy performance in other ways. As an alternative, we investigate how to leverage the detectability of idling behavior to inform exploration and policy improvement.
Our approach, Pause-Induced Perturbations (\ours), applies perturbations at detected idling states, thus helping it to escape problematic basins of attraction. On a range of challenging simulated dual-arm tasks, we find that this simple approach can already noticeably improve test-time performance, with no additional supervision or training. Furthermore, since the robot tends to idle at critical points in a movement, we also find that learning from the resulting episodes leads to better iterative policy improvement compared to prior approaches. Our perturbation strategy also leads to a 15-35\% improvement in absolute success rate on a real-world insertion task that requires complex multi-finger manipulation.

\end{abstract}

\section{Introduction}


Enabling robots to perform dexterous manipulation tasks remains a significant challenge in robot learning.
While better architectures and larger datasets have led to significant improvements, existing approaches are still far from reaching general task mastery. 
In particular, although many failure modes will inevitably be task- and architecture-specific, there are some common patterns that can be observed across a broad range of different settings.

\begin{figure}[t]
    \centering
    \includegraphics[width=1.0\columnwidth]{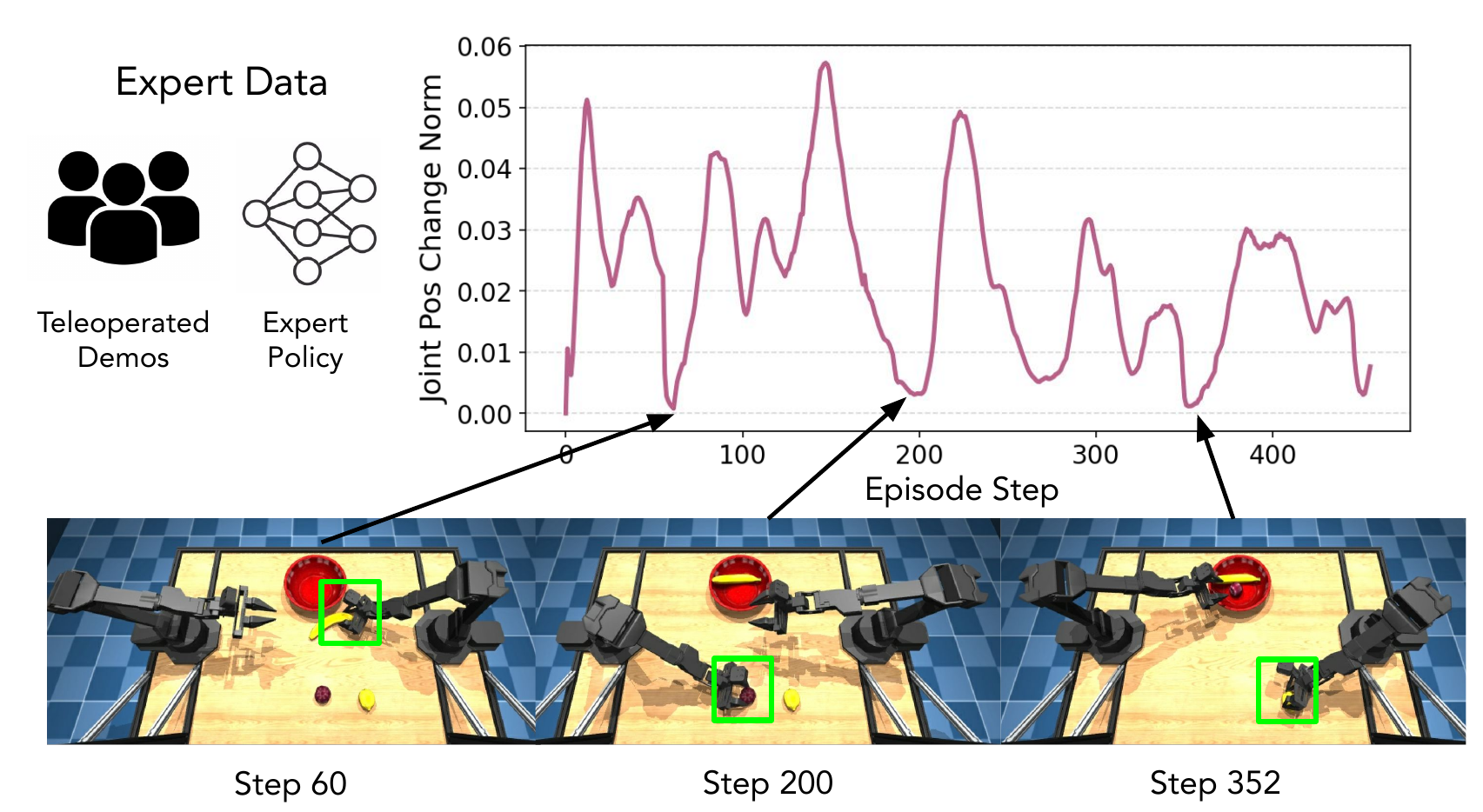}
    \caption{\textbf{Pauses in expert data occur at critical states.} Expert trajectories of dexterous tasks often include subtle pauses, which can lead to idling in policies trained on this data but also provide valuable information about task difficulty. We propose ways of exploiting this indefinite pausing for better performance, as opposed to techniques for removing this behavior.}
    \label{fig:teaser}
\end{figure}

For instance, successful robot episodes, collected either by human demonstrators or with an expert policy, often contain subtle pauses preceding difficult maneuvers, as shown in Figure~\ref{fig:teaser}.
When state-of-the-art imitation learning policies are trained on unfiltered data with these pauses, they exhibit "idling behavior," mirroring these pauses during their own rollouts, as observed in prior work~\cite{zhao2023learning,chi2023diffusion}.
This behavior may also be caused or exacerbated e.g. by uncertainty due to data scarceness, or action multi-modality in certain parts of the state space~\cite{swamy2022causal,liu2024bidirectional}. In our investigation, we observe this behavior as a general phenomenon for different tasks, policy architectures, and learning methods.

While there are ways to reduce the policy idling behavior, e.g., by filtering pauses and small actions from the training data or by executing longer sequences of actions open loop, our experiments suggest that such approaches may require careful tuning and can compromise policy reactivity and precision. 
Moreover, such approaches do not capitalize on the following insight: idling behavior tends to occur before difficult or intricate maneuvers that require small actions, as we observe in failure trajectories in Figure~\ref{fig:rollouts}, and may thus provide a valuable signal for policy improvement.

Our method, Pause-Induced Perturbations (\ours), 
takes a different approach. It identifies idling behavior and leverages it as a signal for exploration and policy improvement.
At test time, we detect "idling behavior" and apply a perturbation to the action commands towards the robot arms' initial position. The concept of using perturbations to escape unwanted attractors is well-studied in fields like physics and chemistry, where it is applied to a variety of dynamical systems~\cite{rodrigues2010escape,kramers1940brownian}. Drawing on this principle, intuitively our perturbations aim to dislodge the policy from the problematic local states associated with idling, encouraging exploration of nearby states and potentially leading to task completion. This has two main benefits: (1)
we find that test-time perturbations offer immediate performance gains, and (2) we can exploit the resulting perturbed rollouts for iterative policy self-improvement.  

Current methods for iterative self-improvement tend to plateau quickly with additional rollouts~\cite{mirchandani2024so}, due to limited exploration in informative regions of the task space. 
Our pause-induced perturbations implicitly focus on challenging segments of the task, and can thus lead to a richer dataset of successful trajectories. 
Building on this insight further, we leverage idling detection 
to pinpoint where mistakes may have occurred in failure trajectories and exploit this in the policy improvement step: 
we fine-tune the policy with preference-based imitation learning using labels obtained from idling detection, to reduce the likelihood of transitions leading to idling segments in failure trajectories.


The main contributions of this work are the following:
\begin{enumerate}
    \item We establish that policy failures, in a variety of settings, are caused by idling behavior.
    \item We propose an autonomous perturbation strategy at detected idling states for test-time improvement.
    \item We show that the perturbed rollouts also aid iterative policy improvement and propose a preference-based imitation learning approach that leverages idling detection to provide informative preference labels, reducing the likelihood of transitions leading to idling in future rollouts. 
\end{enumerate}
We evaluate our approach on a suite of nine challenging simulated ALOHA tasks~\cite{zhao2023learning} and a real-world connector insertion task with a dexterous hand~\cite{bauza2024demostart}.
We find that applying simple perturbations at detected idling states leads to immediate improvements in test-time performance by helping the policy escape these problematic basins of attraction. 
Our test-time pause-induced perturbation strategy, which requires no additional training or supervision, leads to a 5-10\% absolute improvement in success rate in the ALOHA tasks compared to the base policy, and a 15-35\% absolute improvement on the real-world connector insertion task. Furthermore, we show that the knowledge gained through this targeted exploration can be effectively distilled back into the policy, resulting in similar improvements in the next round of iterative policy improvement. 


\section{Related Work}

\subsection{Detecting and recovering from challenging states}
Previous work has explored identifying critical states \cite{spielberg2019concept} or bottlenecks \cite{menache2002q} in a task via methods like sensitivity analysis \cite{liu2023learning} or using heuristics based on task progress~\cite{ecoffet2021first}. Other approaches have learned a critic that can predict the difficulty of a state~\cite{florensa2017reverse}, where the agent might be stuck~\cite{du2024err}, which sampled action to take~\cite{nakamoto2024steering}, or how to recover from out-of-distribution states~\cite{chen2022you}. Other works use density estimation to identify states that are out of distribution and utilize a specialized policy to recover \cite{reichlin2022back}. However, these methods often require significant computational overhead, reliable value functions (which can be hard to learn from offline data) or external information (e.g., expert behaviors). Our method, on the other hand, provides a simple and effective way to identify critical states by directly observing the "idling behavior" exhibited by policies trained on unfiltered expert data. Our method is inspired by the use of perturbations to escape from unwanted attractors in dynamical systems, which has been well-studied in physics and chemistry \cite{rodrigues2010escape,kramers1940brownian}.

\subsection{Self-improvement flywheels}

Iterative policy improvement~\cite{bousmalis2023robocat,ahn2024autort} has demonstrated the potential of bootstrapping performance from initial demonstrations through iterative data collection and policy refinement. Works like DAgger~\cite{ross2011reduction} and its variants~\cite{laskey2017dart,kelly2019hg,hoque2021thriftydagger} rely on expert intervention for data collection, which is less scalable and more expensive. Some recent work lessens this cost but still requires human labels for where to augment the data with perturbed transitions~\cite{ankile2024juicer}. 
This setting is similar to offline-to-online reinforcement learning~\cite{nair2020awac,lee2022offline,nakamoto2024cal} in that one first trains on offline data and subsequently performs policy rollouts to further learning, but with the important difference that no online RL is used, which can be harder to set up efficiently with larger models and can require careful tuning of replay buffer settings. Another important difference is that offline-to-online RL relies on value functions, which are less obvious to learn with action chunking and diffusion policies, although there has been some recent work in that direction \cite{hansen2023idql,seo2025coarsetofineqnetworkactionsequence,ren2024diffusion}. 
Our work aims to improve the self-improvement flywheel of iterative policy improvement by leveraging an intrinsic signal---idling---to guide exploration and overcome performance plateaus.

\subsection{Exploration strategies}

A variety of exploration techniques have been proposed to address the exploration-exploitation dilemma in RL and, by extension, iterative policy improvement. Generic exploration strategies, such as epsilon-greedy or Boltzmann exploration~\cite{sutton1998reinforcement,osband2016deep}, often lack the targeted focus needed for efficient improvement. Curiosity-driven methods~\cite{pathak2017curiosity,still2012information,burda2018large,lambert2022challenges} explore regions of the state space that are novel or unpredictable, but may not directly address the specific challenges faced by the policy in performing the task. 
Another class of methods adds different types of noise into the data collection process. Methods like DART~\cite{laskey2017dart} and Bayesian Disturbance Injection \cite{oh2023bayesian} add noise during demonstrations, whereas other methods~\cite{brown2020better,zisman2023emergence} inject varying amounts of noise into the learned policy. Noise can also be injected into the demonstrations to robustify training~\cite{ke2021grasping,ke2023ccil}.
Our work proposes a data-driven exploration strategy that leverages pauses, which are easily detectable and commonly found in rollouts, to guide targeted exploration specifically in challenging segments of the task.


\section{Preliminaries}

Our goal is to maximize the expected cumulative reward, $\mathbb{E}[R] = \mathbb{E} \left[ \sum_{t = 0}^T \gamma^t r(s_t, a_t) \right]$,
within a Markov Decision Process (MDP) defined by $(\mathcal{S}, \mathcal{A}, \mathcal{P}, r)$ with state space $\mathcal{S}$, action space $\mathcal{A}$, transition probabilities $\mathcal{P}(s'|s, a)$, and reward $r(s, a)$. We are provided with an initial dataset of $N$ successful episodes, 
$D_{\text{expert}} = \{\tau_1, \ldots, \tau_N\}$, where each $\tau_i = \{s_0, a_0, s_1, a_1, \ldots, s_T, a_T\}$ is a state-action trajectory. These trajectories may be provided by humans or other means, for example scripted or expert RL policies. Towards our goal, we learn a policy $\pi_{\theta}(a | s)$ and apply test-time perturbations, with no additional human supervision, model training, or data needed. 

We additionally consider an iterative policy improvement setting, derived from the Collect and Infer paradigm~\cite{riedmiller2022collect,lampe2024mastering}, which leverages a limited online interaction budget to iteratively improve the policy. An initial policy, $\pi_0$, is first trained on $D_{\text{expert}}$. It is then used to collect a dataset of online experience, $D_{\text{aug}} = \{\tau_1, \ldots, \tau_M\}$. Each trajectory $\tau_j$ is associated with a binary success indicator, that captures whether the task was completed successfully. These rollouts are then used to train or fine-tune the next policy, $\pi_{1}$, and this can be repeated for further iterations. A key challenge in iterative policy improvement is balancing exploitation of successful behaviors with exploration to discover potentially superior strategies. The limited online budget necessitates efficient exploration to maximally improve the policy within the allowed number of interactions.  


\section{Pause-Induced Perturbations (\ours)}

\subsection{Detecting idling behavior from the policy}



\begin{figure*}[t]
    \centering
    \includegraphics[width=0.9\textwidth]{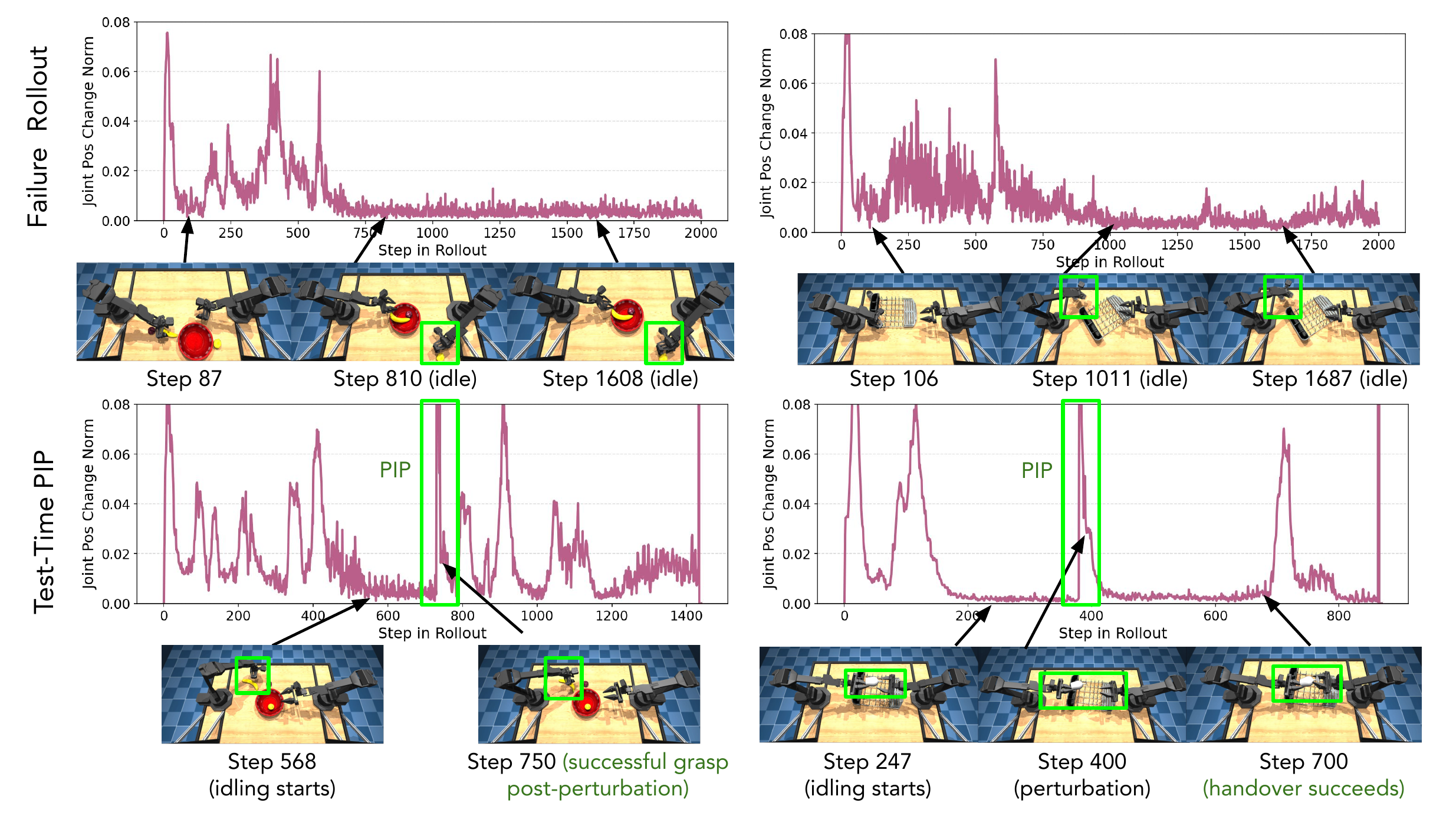}
    \caption{\textbf{Idling behavior in policy rollouts indicates critical states.} 
    (Top Row) Failure rollouts on the Fruit Bowl and Glass on Rack tasks. The plot depicts the $\ell2$-norm of the change in joint position over time, with low values indicating periods of idling. Corresponding images show the robot's configuration during these pauses, occurring before or during challenging maneuvers (e.g., grasping fruit or glass).  (Bottom Row) Test-time execution on the tasks using our perturbation strategy. Corresponding images illustrate how the perturbation helps the robot escape idling states and successfully complete the tasks.
    }
    \label{fig:rollouts}
\end{figure*}

We train an initial policy $\pi_0$ on the expert data $D_{\text{expert}}$ and then execute it in the environment and observe its behavior. We find that $\pi_0$ often exhibits idling behavior characterized by prolonged periods of limited movements, particularly at critical state regions with small actions present in the expert data.
We define idling behavior as a sequence of consecutive states where the $\ell2$-norm of the change in robot joint positions between consecutive transitions falls below a threshold $\epsilon$ for a duration exceeding $T$ timesteps.
Figure~\ref{fig:rollouts} shows examples of this idling behavior, highlighting the correspondence between difficult parts of the task and inaction in the policy's execution.

\subsection{Perturbations and self-improvement using detected idling}

Our approach hinges on the observation that pauses and small actions in expert data often correspond to critical states requiring precision in dexterous manipulation tasks, and that this is also reflected in policy idling. 
We thus leverage idling behavior for both test-time performance improvement and targeted data collection for policy self-improvement. 

\paragraph{Test-time perturbations}
During test-time rollouts of $\pi_0$, we detect "idling states" where the policy exhibits indefinite pause behavior (as defined above). Upon detecting an idle state, we take a perturbed action $a_{\delta}$. This perturbation is a simple alteration to the predicted action, by interpolating the action towards the initial joint configuration of the robot arms. 
That is, if the policy is position-controlled, instead of executing the predicted action, we take the following action interpolating between the current joint positions $s_{\text{joints}, t}$ and the initial joint positions $s_{\text{joints}, 0}$:
\[a_{\delta} = \sigma s_{\text{joints}, t} + (1 - \sigma) s_{\text{joints}, 0} \, , 
\]
where $\sigma$ is a hyperparameter controlling the perturbation magnitude and balancing exploration and exploitation. Empirically, we find that a value of $\sigma$ between 0.6 and 1.0 works well; this keeps the perturbed action closer to the original prediction, breaking the idling behavior while still leveraging the policy's knowledge. Intuitively, this perturbation to the executed actions aims to dislodge the policy from the local states associated with idling behavior, encouraging exploration of nearby states and potentially leading to task completion.

\paragraph{Iterative policy improvement with preference-based imitation learning}

For autonomous policy improvement, we collect the trajectories resulting from these test-time perturbations, including both successful and unsuccessful outcomes, to form an augmented dataset $D_{\text{aug}}$.
Crucially, idling detection provides valuable information for refining the policy---by identifying idling segments within failure trajectories, we can pinpoint actions that likely contributed to failure.  We leverage this insight within a preference-based learning framework, PMPO~\cite{abdolmaleki2024preference}, to fine-tune the policy.

Specifically, we maximize the likelihood of all transitions from successful trajectories (denoted as $D_s$) and minimize the likelihood of transitions leading to idling states in failed trajectories (denoted as $D_f$) and a KL-divergence term to regularize the updated policy towards the initial policy $\pi_0$, to prevent excessive deviation:

\begin{align*}
L_{\text{PMPO}}(\theta) = &\alpha \mathbb{E}_{(s, a) \sim D_s}[\log \pi_\theta(a|s)] \\
&- (1-\alpha) \mathbb{E}_{(s, a) \sim D_f}[\log \pi_\theta(a|s)] \\
&- \beta \text{KL}(\pi_{\text{ref}}, \pi_\theta; s)\, ,
\end{align*}
where $s$ denotes the state, $a$ denotes the action, $\alpha$ controls the weighting between accepted and rejected actions, and $\beta$ regulates the KL divergence penalty. 
\section{Experiments}
\label{sec:exps}

In this section, we first describe our experimental setup and analyze the prevalence of policy idling. We then empirically evaluate (1) the extent to which \ours improves the test-time performance of the base policy in both simulated and real-world settings, and (2) if \ours can lead to better iterative policy improvement compared to prior methods. 

\begin{figure*}[t]
    \centering
    \includegraphics[width=1.0\textwidth]{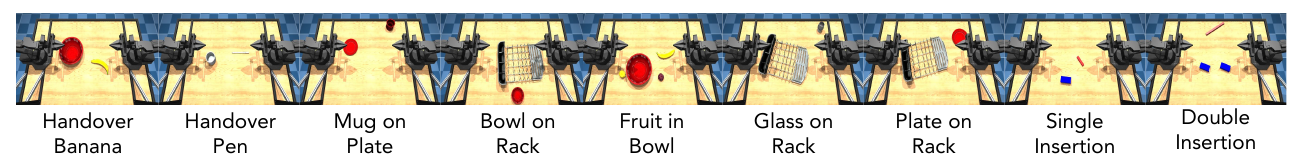}
    \caption{\textbf{Simulated ALOHA test suite.} We evaluate on nine challenging bimanual tasks in a simulated ALOHA environment, encompassing a range of objects (e.g. mug, fruit, glass, peg) and initial states that often require precise manipulation to complete successfully.}
    \label{fig:tasks}
\end{figure*}

\subsection{Experimental Setup}

\subsubsection{ALOHA simulation task suite}
We test \ours on a range of nine challenging bimanual tasks in an ALOHA~\cite{zhao2023learning} environment simulated in MuJoCo, shown in Figure~\ref{fig:tasks}.
The observation space consists of joint positions, an overhead camera, worms-eye camera, and 2 wrist cameras.
The action space consists of 14 joint positions including the gripper. The control rate is 50Hz.
Each of these tasks have a wide variety of initial states. Demonstrations are collected via human teleoperation using an Oculus headset. For each task, we begin with 214 to 616 human-collected demonstrations as expert data. We evaluate and report success rates and 90\% confidence intervals over 1000 trials, with a maximum episode length of 2500 steps. 

In this domain, we mainly use the Perceiver Actor Critic (PAC) architecture~\cite{springenberg2024offline} together with a SigLIP vision encoder~\cite{tschannen2025siglip} as the foundation for our policy representation. 
The perception module of PAC encodes these inputs into a compact latent representation with a cross-attention encoder.
The policy network is implemented using a cross-attention decoder to predict the parameters of the action distribution from the encoder's latent token representation.
We model the action distribution using a Laplace distribution, with its parameters (mean and scale) outputted by the actor network.
Minimizing the negative log-likelihood of a dataset of samples under this distribution is equivalent to minimizing an $\ell1$ regression loss on the same data, which has been shown to be effective in prior work in the ALOHA domain~\cite{zhao2023learning}, and we found it to be empirically better than using Gaussian action distributions.
We train with action chunking~\cite{zhao2023learning}, using a training chunk length of 50 and open loop execution of 20 steps during evaluation. For \ours, we detect idling if the total joint pose change is under the threshold $\epsilon = 0.06$ across adjacent size-20 chunks for $T = 8$ timesteps, setting $\sigma = 0.6$ for the perturbation magnitude. These hyperparameters were chosen based on preliminary experiments, were held constant for all tasks, and were not extensively tuned.

To ensure our conclusions are not specific to a particular policy class, we also evaluate the test-time performance improvement with \ours for a multi-task diffusion policy with an architecture used in~\cite{zhao2024aloha} on a subset of the tasks, with the same default hyperparameters and action chunking of 50 steps with 10 open loop steps.

\subsubsection{Real-world connector insertion with a DEX-EE hand}

\begin{figure}[t]
    \centering
    \includegraphics[width=1.0\columnwidth]{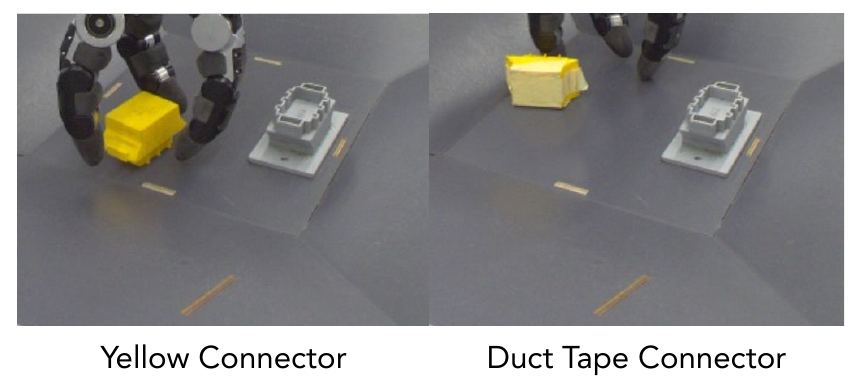}
    \caption{\textbf{DEX-EE connector insertion task.} We evaluate on a precise real-world dexterous manipulation task, where a three-fingered DEX-EE hand must pick up the connector and fit it into the socket.}
    \label{fig:dextask}
    \vspace{-3mm}
\end{figure}

We additionally evaluate the test-time performance improvement of \ours on a dexterous real-world task. 
We follow the experimental setup used in DemoStart~\cite{bauza2024demostart}, consisting of a Kuka LBR iiwa 14 robot arm equipped with a three-fingered DEX-EE hand, with dual wrist-mounted cameras positioned at the hand's base and tactile sensors at the fingertips. The 18-dimensional action space includes 6 dimensions for the hand's desired Cartesian velocity and 12 dimensions for the desired joint angles of the hand's fingers. We evaluate using 20 trials of a Connector Insertion task with two connector variants, one yellow and one covered with duct tape with higher friction for slightly easier grasps. We first train a policy with privileged state information using DemoStart in a MuJoCo simulation environment, and then distill expert episodes to a student policy with visual observations using ACT~\cite{zhao2023learning} for the real world. The distillation dataset consists of 1 million expert trajectories collected with physics domain randomization on the friction, mass, and inertia of the scene joints and bodies, as well as applied force variation for the free bodies in the scene. 
We randomize asset colors and textures for 700k trajectories and use photorealistic Filament renderings with randomized lighting for the remaining 300k. The ACT policy receives five camera images (three overhead and two wrist), arm and hand joint angles, wrist pose, and binary fingertip contacts. 

Transferring from simulation to the real world involves a significant distribution shift, which frequently induces idling behavior. We detect such behavior in the real world by thresholding the average difference between consecutive action predictions using a threshold $\epsilon = 0.001$ for $T = 2$ timesteps. After a pause is detected, the perturbation moves the arm up by 5cm (towards the initial robot arm position) over 10 timesteps (at a control rate of 20Hz), then resume policy execution. 

\subsection{Baselines}
To evaluate the test-time performance improvement of \ours in the simulated ALOHA environment, we compare to the following methods:
(1) Base, where we train a base policy on all the given expert data and successful online rollout episodes, (2) Noise~\cite{sutton1998reinforcement}, where we add Gaussian noise to the actions in an $\epsilon$-greedy fashion, (3) RND~\cite{burda2018exploration}, where we sample multiple actions at each timestep and choose one based on their novelty measured by the RND score computed over the trajectory of (proprioceptive state, action) pairs observed during the current episode.
We additionally compare performance to a base policy trained with small actions filtered out and with a longer action chunking sequence (Pause-Filtered), which are strategies prior work has used to eliminate pausing~\cite{zhao2023learning}.

During the iterative improvement step, we fine-tune a base policy with the rollouts collected using the above methods, using the standard iterative improvement recipe of fine-tuning with behavioral cloning on only successful rollouts. For \ours, we use $\alpha = 0.9$ and $\beta = 0.3$ for the preference-based imitation learning loss.

\begin{figure*}[t]
    \centering
    \includegraphics[width=1.0\textwidth]{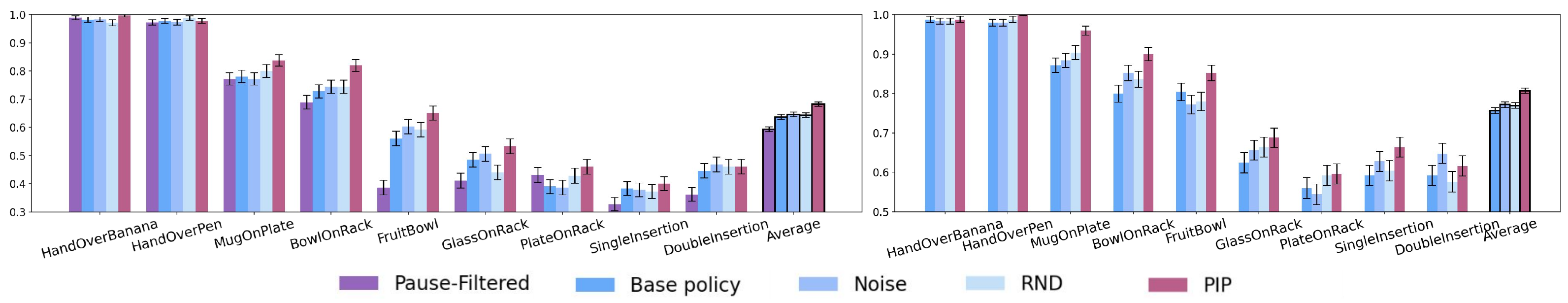}
    \caption{\textbf{Test-time perturbations with \ours on ALOHA tasks with PAC agents.} For both (left) a base policy trained only on demos and (right) one trained on both demos and successful rollouts from 5000 trials per task, \ours significantly improves performance over the base policy and comparisons with no additional training or supervision. We report success rates over 1000 trials and 90\% confidence intervals for each task. }
    \label{fig:results}
\end{figure*}


\begin{figure}[t]
    \centering
    \includegraphics[width=0.7\columnwidth]{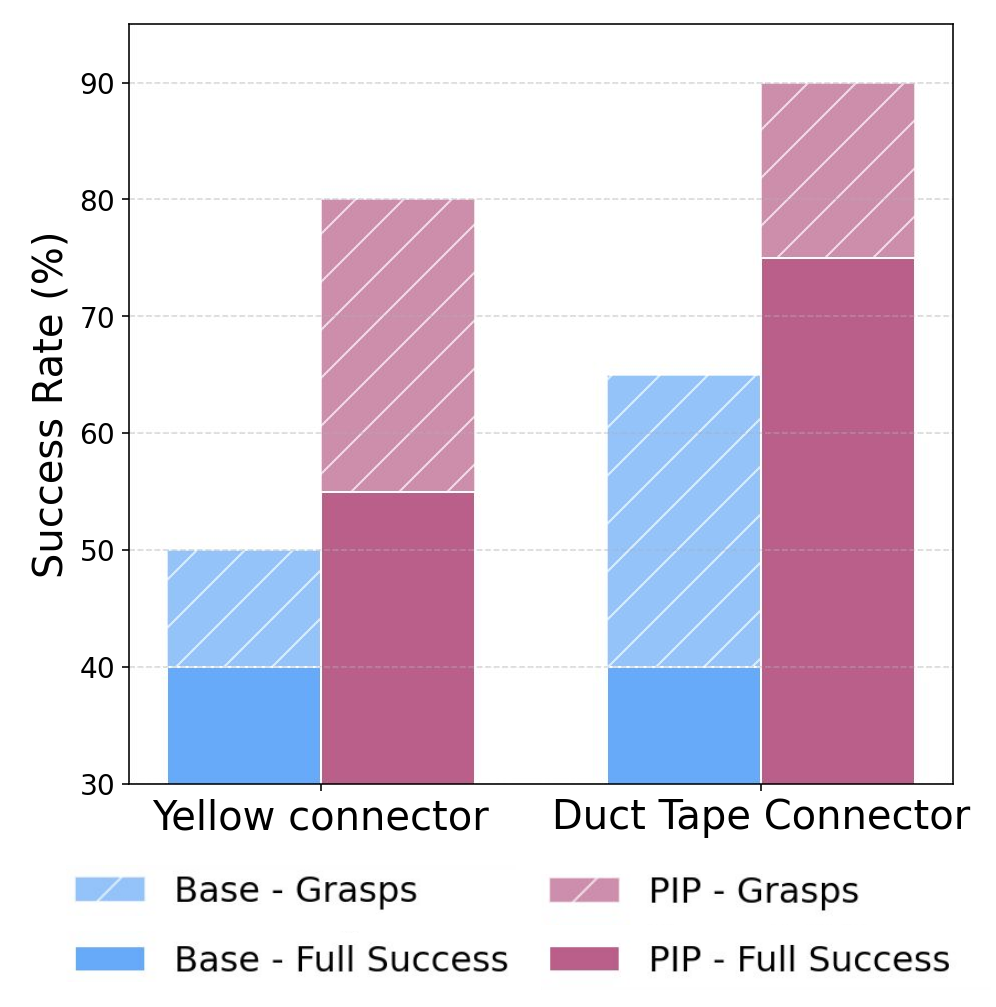}
    \caption{\textbf{\ours significantly improves performance on a real-world DEX connector insertion task.} We report success rate across 20 trials for each method on an insertion task with two different connectors and find that perturbations with \ours improve performance by 15-35\% absolute success rate over the base policy on grasping the connector and full task success.}
    \label{fig:dex_results}
    \vspace{-4mm}
\end{figure}

\subsection{Policy idling failure mode analysis}


\begin{table}[t]
    \centering
\begin{tabularx}{0.95\columnwidth}{c|X|c}
\toprule
\textbf{Domain} & \textbf{Method} & \textbf{Average \% Failures Idle} \\ 
\midrule
ALOHA & PAC (Filter + High OL) & 3.5\% \\ 
 & PAC & 84.9\% \\ 
 & Diffusion & 22.3\%\\
 \midrule
DEX-EE  & ACT & 90\% \\ 
\bottomrule
\end{tabularx}
    \caption{\textbf{Average percentage of failures with idling.} A high percentage of failures contain idling across our settings and policies, showing that this is a significant failure mode for both simulated and real-world settings. We can avoid pauses by filtering out small actions from the dataset and increasing the number of open-loop steps (Filter + High OL), but this can hurt performance, as shown in Figure~\ref{fig:results}. }
    \label{tab:pauses}
\end{table}

\begin{table}[t]
    \centering
\begin{tabularx}{0.8\columnwidth}{c|c}
\toprule
\textbf{Method} & \textbf{Average Action Variance} \\
\midrule
PAC (Filter + High OL) & 2.7e-3 \\ 
PAC  & 2.2e-3 \\ \midrule
PAC, idle states & 3.8e-4 \\ 
PAC, non-idle states & 2.3e-3 \\
\bottomrule
\end{tabularx}
\caption{\textbf{Action variance in idle versus non-idle states.} We compare the average variance of actions taken by PAC policies across different conditions. The lower action variance in idling states compared to non-idling states for the base policy indicates that the policy predicts small, repetitive movements reflecting the small-action training data in critical states.}
\label{tab:variance}
\vspace{-4mm}
\end{table}


Idling behavior often arises when the robot, operating near states that require precise actions, becomes stuck predicting and executing small, repetitive movements. To understand the prevalence of policy idling as a failure mode, we analyzed the frequency of failure episodes that contain idling in our dexterous manipulation settings, shown in Table~\ref{tab:pauses}. In our main evaluation on the ALOHA test suite, for the PAC policy trained on unfiltered demonstrations, a high 84.9\% of failed episodes contained idling behavior. Similarly, in the real-world DEX Connector insertion task, idling occurred for 90\% of failures with the base policy. The incidence of idling was lower for the diffusion policy but was still a significant failure mode. As a comparative baseline, we trained a PAC policy on a dataset where small actions (with $\ell2$-norm of consecutive actions under 0.0015) were filtered out, and used a higher open-loop setting of 40 during execution (Filter + High OL). This approach drastically reduced the incidence of idling in failures in ALOHA down to 3.5\%. However, as seen in Figure~\ref{fig:results}, mitigating idling through filtering and high open-loop control can compromise overall task performance, which motivates our approach to leverage rather than eliminate these transitions with small actions.  

We investigate two potential causes of idling behavior: (1) the policy repeatedly predicts the same suboptimal actions, or (2) the policy exhibits high-variance action predictions, rapidly switching between different actions. To assess these hypotheses, we calculated the average variance of action samples at both idling and non-idling states for the base policy and a non-idling policy (Filter + High OL). As shown in Table~\ref{tab:variance}, idling states exhibited significantly lower average action variance than non-idling states for the base policy.  Furthermore, the non-idling policy showed even higher action variance. These findings support the first hypothesis, suggesting that idling arises from the policy consistently predicting the same, ineffective action in specific state regions, rather than from high-variance action switching.


\begin{figure*}[t]
    \centering
    \includegraphics[width=1.0\textwidth]{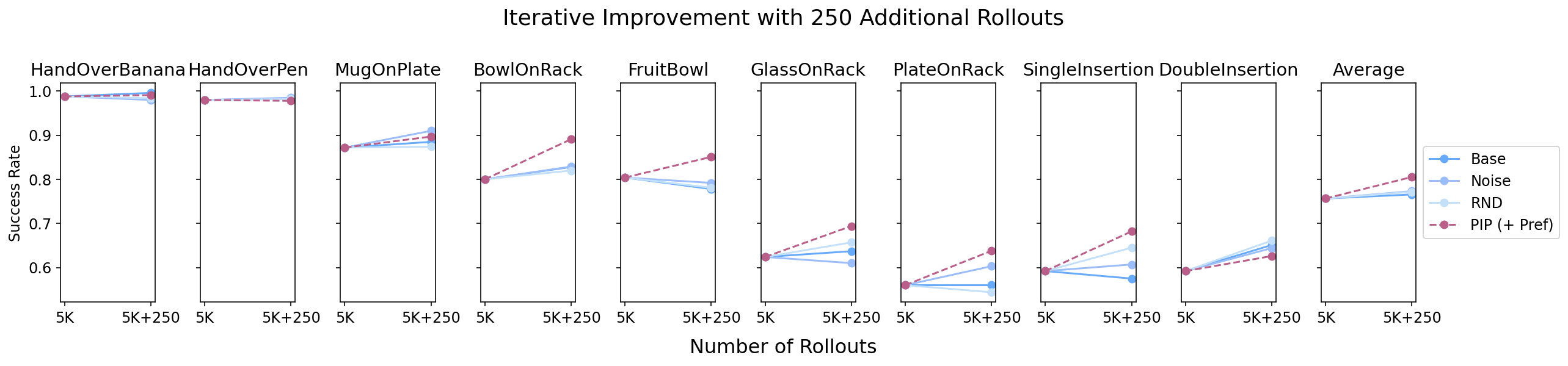}
    \caption{\textbf{Iterative improvement plateaus without \ours.} After training a base policy with demos and 5000 rollouts generated by an initial policy trained on demos, iterative imitation learning, even with noise or RND exploration, shows negligible improvement with an additional 250 rollouts, indicating a plateau in performance. However, \ours (+ Pref) continues to improve, highlighting the benefits of targeted exploration at critical states.}
    \label{fig:postp}
\end{figure*}

\begin{figure*}[t]
    \centering
    \includegraphics[width=1.0\textwidth]{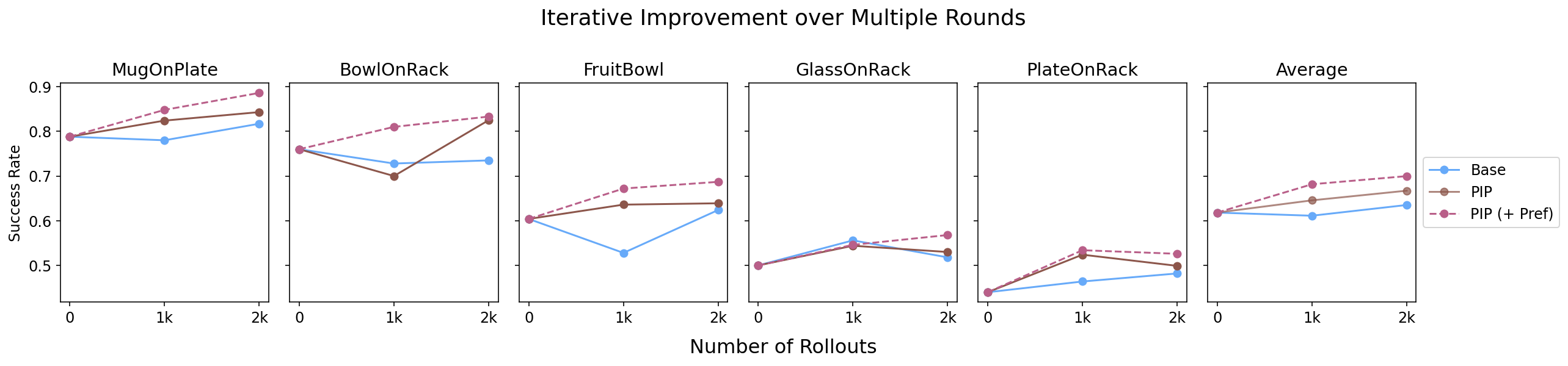}
    \caption{\textbf{\ours improves over multiple rounds of iterative improvement.} Across two rounds of iterative policy improvement each with 1000 rollouts per task, \ours (+ Pref) consistently outperforms the base policy and a variant without preference-based training.}
    \label{fig:mult}
\end{figure*}

\subsection{Improving test-time performance}

Our test-time perturbation strategy consistently demonstrates improved performance across both simulated and real-world dexterous manipulation tasks. In the following, we show how \ours can exploit idling in three policy and environment settings: simulated ALOHA with PAC, simulated ALOHA with diffusion policies, and real-world DEX-EE with ACT.

In the simulated ALOHA test suite, as shown in Figure~\ref{fig:results}, applying perturbations at detected idling states resulted in a 4.4\% overall average increase in absolute success rate compared to a PAC policy, with no additional human supervision, for two different base policies (one trained only with demonstrations and another trained additionally with the successful rollouts from 5000 trials per task). We observe that the benefits are most noticeable in tasks such as Fruit Bowl, where  PIP improves performance by almost 10\%, and the Object On Rack tasks, where PIP improves performance by an average of 7\%. In these scenarios, the perturbations provided by \ours are often sufficient to overcome idling, allowing the policy to achieve success where it would previously pause. The gains are less significant in the insertion tasks---in these tasks, the objects often end up in significantly out-of-distribution configurations, where local perturbations are less likely to bridge the gap to states similar enough to a successful trajectory from the training data. 

As a baseline, we tried several approaches for filtering out small actions with various thresholds, along with a longer open loop of 40 timesteps to remove idling. However, even with extensive tuning, these often degraded the overall performance of the policy; we report the performance of best model with filtering as Pause-Filtered. The reason for the suboptimal performance may be because small actions often coincide with difficult parts of the task requiring precision, so filtering out even a small amount of these actions may lead to reduced precision. \ours also consistently outperformed common perturbation methods that either add noise to actions or utilize RND for exploration, highlighting the value of having a targeted strategy for perturbations when the policy is idle.


\begin{table}[t]
    \centering
\begin{tabularx}{0.8\columnwidth}{X|X|c}
\toprule
\textbf{Task} & \textbf{Method} & \textbf{Success Rate} \\ 
\midrule
PlateOnRack & Base & 46.7\% \\ 
 & \ours & 56.7\% \\ 
 \midrule
FruitBowl & Base & 58.3\% \\ 
 & \ours & 63.3\% \\ 
\bottomrule
\end{tabularx}
    \caption{\textbf{PIP improves test-time performance using a diffusion policy.} \ours shows significant improvement in success rate on two simulated ALOHA tasks with this policy class, across 60 trials.}
    \label{tab:diff}
    \vspace{-4mm}
\end{table}

Furthermore, to demonstrate the potential applicability of our method to different policy architectures, we also apply \ours to a diffusion policy following~\cite{zhao2024aloha}, trained on the Plate On Rack and Single Insertion tasks. While the behavior of this policy exhibits more consistent movement due to better handling of multimodality~\cite{chi2023diffusion}, the policy can still become stuck retrying the same strategy in states near a critical state region, and we can detect this and leverage \ours by using a higher threshold ($\epsilon = 0.02$ across adjacent actions for 200 steps). In Table~\ref{tab:diff}, we find that \ours can still lead to improvements in test-time performance in this setting.

Our perturbation strategy shows particularly significant improvements in the real-world DEX connector insertion task across two different connectors, as shown in Figure~\ref{fig:dex_results}. \ours boosted grasp success by 25-30\% and full insertion by 15-35\%, in terms of absolute success percentage. 
While the robot may initially pause due to sim-to-real distribution shift or partial observability in the real world, where the policy cannot distinguish small differences between neighboring states, our perturbations are often sufficient to resume progress, enabling the policy to successfully complete the task.
Note that in this case there were no pauses in the training data, since it was trained on expert RL agent data, so filtering small actions was not an option.

These results highlight that policy idling can originate from different sources: imperfections in human demonstrations and the sim-to-real gap. In both cases, the policy can become trapped in states characterized by small, repetitive actions, ultimately leading to idling and task failure. By recognizing and leveraging this shared characteristic, our perturbation strategy offers a unified approach to improving performance in both demonstration learning and sim-to-real transfer settings.

\subsection{Better iterative policy improvement data}

A key challenge in iterative policy improvement is the tendency for performance improvement to plateau after training with policy rollouts~\cite{mirchandani2024so}. We see this in Figure \ref{fig:postp}: after an initial round of training on 5000 environment rollouts, collecting and training on an additional 250 rollouts provides negligible benefit to the baseline policy across the ALOHA task suite.  This suggests that these additional rollouts do not provide further useful information. In contrast, \ours (+ Pref) continues to improve even with this limited amount of additional data, demonstrating the effectiveness of our targeted exploration strategy on states identified through pause detection. 

To further investigate the benefits of our approach, we conducted experiments with multiple rounds of iterative improvement, as shown in Figure \ref{fig:mult}. In this setting, policies were trained on 1000 rollouts per task for two rounds. We find some improvement with incorporating preference information derived from pauses, as \ours (+ Pref) improves by 3\% average success rate across all tasks, compared to a variant of \ours that does filtered behavior cloning and does not utilize the preference-based training objective, showing that decreasing the likelihood of actions leading to pauses is another way of leveraging the detected pauses. Furthermore, \ours (+ Pref) exhibits consistent and significant improvement across both rounds compared to the base policy, highlighting the sustained benefit of our targeted exploration strategy.


\section{Limitations and Future Work}

While our method demonstrates the effectiveness of leveraging idling in policy rollouts for improved dexterous manipulation, several limitations and avenues for future work exist.
First, our current perturbation strategy employs a simple interpolation towards the initial joint configuration. If the tasks have a brittle training distribution where the robot is likely to end up in out-of-distribution states, the policy may not always be able to recover even with perturbations towards more in-distribution states. Exploring more sophisticated perturbation techniques could potentially yield further performance gains.  
In addition, \ours may not lead to significant improvements for policies where idling is not a main failure mode, such as expert RL policies or environments with abundant data. 
Finally, strategies to overcome and leverage policy idling open up numerous exciting avenues for future research. For instance, the detected idling states could inform the selection of new expert demonstrations, guiding data collection towards the most challenging aspects of the task.  Integrating pause detection mechanisms with interactive learning approaches like DAgger could also enable more targeted human interventions, further accelerating the learning process. 




\bibliographystyle{IEEEtran}
\bibliography{example}



\end{document}